\documentclass[runningheads]{llncs}
\usepackage[T1]{fontenc}
\usepackage{graphicx}
\usepackage{booktabs}
\usepackage{listings}
\usepackage{pifont}
\usepackage{microtype}
\usepackage{hyperref}
\usepackage{hyphenat}

\begin{document}

\title{Agyn: An Open-Source Platform for AI Agents with Scalable On-Demand Execution, Agent Definition as a Code, and Zero-Trust Access}

\titlerunning{Agyn: A Scalable Open-Source Platform for AI Agents Runtime} 

\author{Nikita Benkovich \and
Vitalii Valkov}
\authorrunning{N. Benkovich, V. Valkov}
\institute{
Agyn, Inc.; Mila AI e-Lab\\
\email{\{benkovich,vitalii\}@agyn.io}}
\maketitle              
\begin{abstract}
As organizations move toward production deployments of AI agents, which execute non‑deterministic workflows, maintain stateful sessions, and often operate with privileged access to internal services, the engineering challenge shifts from building individual agents to operating them at scale with proper isolation, governance, and security. In this paper we present Agyn, an open‑source platform designed around three key principles tailored for agent workloads: a signal‑driven, stateful serverless runtime on Kubernetes; a Terraform provider for agent and harness definition; and a security model grounded in zero‑trust and least‑privilege principles. Agyn is agent‑agnostic, model‑agnostic, and cloud‑agnostic.

\keywords{Agentic AI Systems \and AI Agent Orchestration  \and Serverless \and Infrastructure-as-Code \and  Zero‑trust Networking.}
\end{abstract}
%
%

\section{Introduction}

Recent advances in tool-augmented large language models (LLMs) have enabled agentic systems capable of autonomous planning, action execution, and communication~\cite{beyondpipelines2025}. As these capabilities mature, organizations are increasingly deploying large numbers of task-specific agents that operate as first-class organizational identities with access to internal enterprise systems. Consequently, the central challenge is shifting from constructing individual agents to operating large populations of such agents reliably and securely at organizational scale.

A desirable deployment model is emerging in which agent definitions are centrally managed and shared across teams, while each running instance is bound to a single user and a single task. Under this model, a single agent definition may spawn many concurrent short-lived instances. 
This pattern --- many agents, many short lived
instances per user, and ambient access to internal services --- exposes three crucial requirements that existing infrastructure cannot satisfy simultaneously:

\begin{enumerate}
  \item \textbf{Efficient on-demand execution.} 
It is not practical to reserve dedicated computing resources for each agent definition ahead of time. Doing so would lead to large amounts of idle compute that are paid for but never used. Instead, the platform must allocate compute only when an agent actually runs, and immediately reclaim any compute that becomes idle so that it can be reused for other tasks.

  \item \textbf{Agent definition as a code.} 
Agent definitions (prompt, tools, secrets,
resource bounds) reflect important operational decisions with production impact. 
So, the teams should manage
agents with the same governance and safety practices they already use for
the rest of their infrastructure. Namely, the definitions shall be subject to version control, and undergo peer review before changes are accepted and applied.
  
  \item \textbf{Zero-trust access to internal services.} Access must be per-identity, deny-by-default, and credentials must never be exposed to the model.
The platform must not assume that the agent or the AI model is trustworthy just because it runs inside the network. Instead, every request to an internal service (such as a database, an API, or a message queue) must be authorized individually.  This prevents the model from leaking credentials or misusing them, even if it behaves unexpectedly or is compromised.
        
\end{enumerate}

Existing systems address these requirements only partially. Serverless platforms such as Knative and AWS Lambda provide elastic execution but lack native support for stateful conversational sessions and agent-oriented triggers. Infrastructure-as-Code (IaC) frameworks provide declarative resource management but lack abstractions tailored to agent workloads. Zero-trust networking systems require manual identity lifecycle management that does not align with ephemeral agent execution. Consequently, no existing platform provides an integrated solution for all three requirements.

This paper presents {\tt Agyn}~\cite{Agyn_GitHub}, an open-source platform for operating organizational scale agentic systems. In {\tt Agyn}, each agent is defined once as a reusable declarative configuration encapsulating prompts, tools, secret bindings, resource limits, operational policies, etc. These definitions are centrally managed and shared across teams, while each execution instance remains isolated and bound to a single user and task (e.g., a conversational thread). The same agent definition may give rise to many such instances concurrently, each with its own state, identity, and life cycle, isolated from all others.

{\tt Agyn} combines: (1) a signal-driven stateful serverless runtime built on Kubernetes, (2) a Terraform provider for agent and harness management, and (3) a zero-trust security architecture based on least-privilege principles. The platform is agent‑agnostic, model‑agnostic, and cloud‑agnostic.

The remainder of this paper presents the background and motivation, system architecture, design decisions, implementation details, limitations, related work, and concluding remarks.

\begin{figure}[!ht]
   \centering      
   \includegraphics[width=0.9\textwidth]{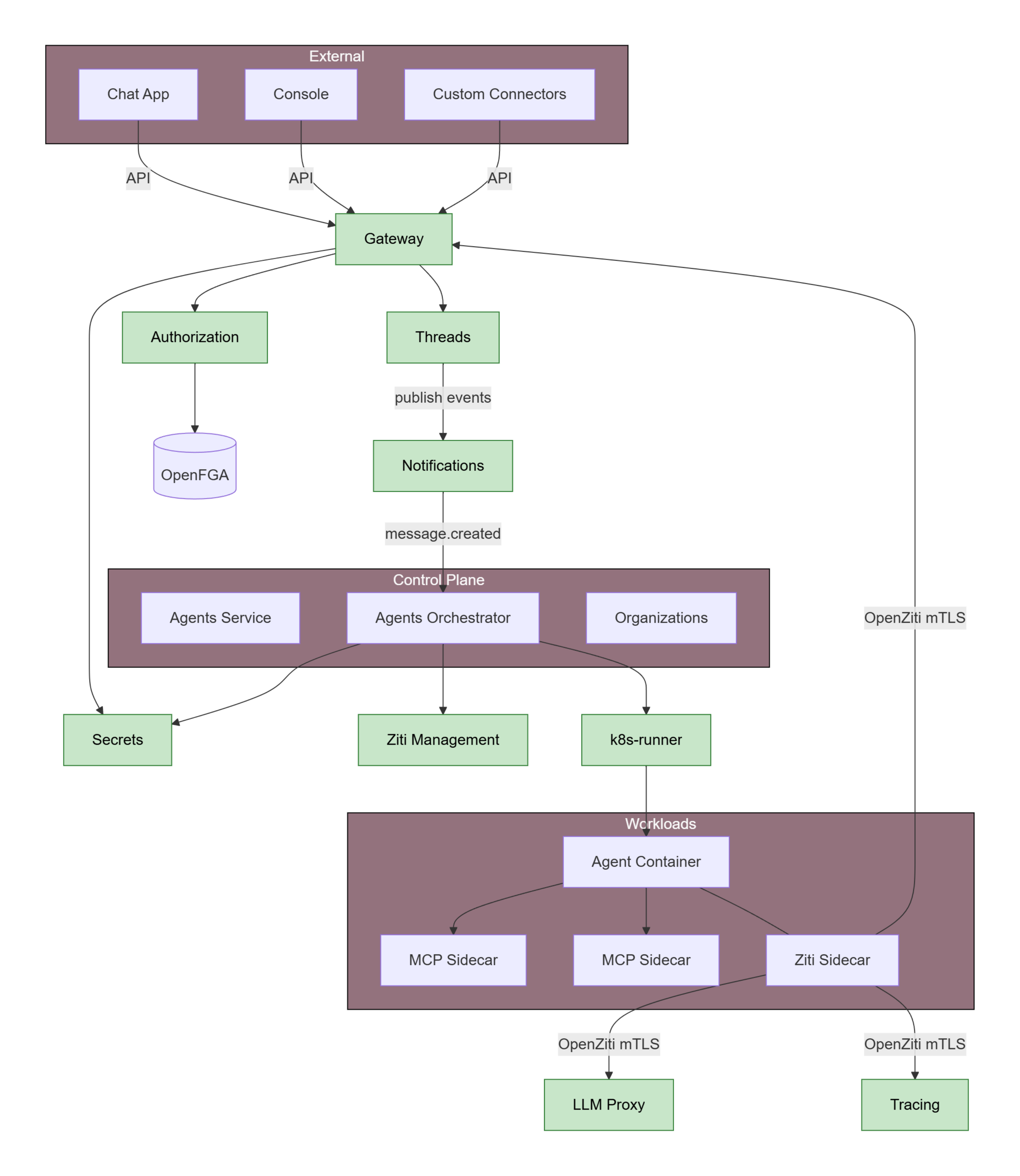}      
 \caption{{\bf Agyn system architecture.} External clients (web chat, console, custom connectors) interact with the platform via the Gateway. The control plane manages agent definitions, orchestration, and tenant state. Agent pods communicate with platform services exclusively through the OpenZiti overlay network (mTLS).
 The data plane elements (filled in green) handle runtime events and agent pod life cycle.
 }
 \label{fig:architecture}
\end{figure} 

\section{Architecture Overview}
\label{sec:architecture}

Agyn is a platform for running AI agents that collaborate with humans
through natural conversation. Users interact with agents via
\emph{threads}: creating a new thread with an agent spawns a dedicated
instance for that user and task. Threads can be created through the
built-in web chat interface or via the platform API, which allows
building custom connectors to external chat applications (Slack,
Teams, etc.). Agents can also create threads with other agents, enabling
multi-agent collaboration. A web console provides administrators with
organisation configuration, access and secret management, and usage
observability.

The platform's core functionality is organised into three areas,
detailed in the following subsections.

\subsection{Runtime: Signal-Driven Stateful Serverless Agents}
\label{sec:runtime}

A naive deployment assigns one long-running worker per agent
definition, wasting compute on idle definitions. Assigning one per
instance is impossible because instance count is user-driven and
unbounded. Traditional FaaS solves scale-to-zero for HTTP
workloads~\cite{jonas2019cloud,hellerstein2019serverless} but the agent
workload differs: the trigger is a conversational message, the unit of
work is multi-minute, and state must survive across spawns. Agyn treats
every instance as ephemeral compute with durable state: spawned on
demand, kept alive while active, reclaimed when idle, enabling efficient
horizontal scaling without pre-provisioning.

\paragraph{Signal-driven spawn.}
When a message arrives on a thread, the Notifications service publishes
an event (Fig.~\ref{fig:architecture}). The Agents Orchestrator
subscribes to these events, determines which agent needs to be running
for that thread, and reconciles: it resolves the agent's secrets,
requests an OpenZiti identity, and instructs the
k8s-runner to create a Kubernetes Pod. The pod mounts the agent's persistent volume (preserving
workspace state from prior sessions) and receives the thread context, so
the agent resumes where it left off rather than starting fresh.

\paragraph{Idle reclamation.}
While active, the agent emits keep-alives every 10\,s. When the idle
timeout elapses (default 5\,min), the Orchestrator stops the pod and
deletes its network identity. The persistent volume and thread history
remain intact; the next message spawns a new instance with the same
state reattached.

\subsection{Configuration: Agents and Harnesses as Code}
\label{sec:configuration}

Once an agent has production responsibility, its prompt, tools, secrets,
and resource bounds are operational decisions that should be reviewable,
versionable, and rolled back like any other
infrastructure~\cite{rahman2019gitops}. Managed platforms expose these
through vendor consoles, invisible to code review and tied to one
vendor. Agyn exposes the entire agent definition and its \emph{harness}
(the surrounding infrastructure reattached on every spawn: MCP servers,
volumes, secrets, network identity) through a Terraform
provider\footnote{\url{https://cf-registry.tf-registry-prod-use1.terraform.io/providers/agynio/agyn/latest}}~\cite{terraform}.

\paragraph{Provider.}
The provider declares the agent and its full harness: agent container,
workspace configuration, system prompt, model, secrets, MCP servers,
persistent volumes, and others.
The platform ships three pre-built agent images (Claude
Code~\cite{claudecode}, OpenAI Codex~\cite{openaicodex}, and Agyn's own agent-loop
implementation); organisations can also bring any custom container.
Terraform's module structure allows organisations to package common
harness patterns (e.g.\ an MCP with its secrets) as shared modules
reusable across agents, avoiding duplication and ensuring consistency.
On \texttt{terraform apply}, definitions resolve through the Gateway
into the Agents Service (see Fig.~\ref{fig:architecture}); subsequent
messages spawn pods from the updated definition without a restart.

\subsection{Security: Zero-Trust and Least Privilege}
\label{sec:security}

Agents must reach internal systems that enforce identity-aware access.
The cluster's service mesh is too coarse, and tunnelling credentials
through the LLM context is unsafe given indirect prompt
injection~\cite{greshake2023not,agenticsecurity2025}. {\tt Agyn} enforces
zero-trust and least privilege through three composing mechanisms.

\paragraph{Per-container isolation.}
Every agent pod's main container hosts the agent process; sidecar
containers host MCP servers~\cite{mcp2024spec}. Each has a separate
filesystem and process tree; only the loopback interface is shared. A
compromised MCP has no view of the agent or other
MCPs~\cite{isolategpt2025}. Compute isolation is a runner-config concern, not part of the agent
definition. Secrets are injected only
into the container that needs them (typically the MCP sidecar), never
into the agent container whose process drives the LLM.

\paragraph{Zero-trust overlay.}
Agents need to reach platform services without exposing credentials to
the LLM. {\tt Agyn} solves this with OpenZiti~\cite{openziti}: every agent
receives its own x509 identity at spawn, establishing who is calling at
the mTLS handshake before application code runs
(Fig.~\ref{fig:architecture}). The Gateway extracts this identity and
propagates it through the call chain, so every downstream service knows
which agent is making the request. ABAC policies then restrict each
agent to only the services it needs; no agent can reach anything not
explicitly permitted.

\paragraph{Relationship-based authorisation (ReBAC).}
Above the network layer, OpenFGA~\cite{openfga} (inspired by
Zanzibar~\cite{zanzibar2019}) enforces fine-grained permissions via
graph traversal. Per-agent roles (owner, maintainer, participant) scope
who can configure or interact with each agent; thread access is
participant-scoped.

\subsection{Implementation and Limitations}
\label{sec:eval}

\paragraph{Technologies.}
Platform services are written in Go. PostgreSQL serves as the sole
durable store; Redis provides pub/sub for event fan-out. The OpenZiti Go
SDK is embedded directly in infrastructure services (avoiding Istio
sidecar conflicts); agent pods use a Ziti TPROXY sidecar for transparent
overlay access. Three identity lifecycle patterns coexist: ephemeral
per-workload agent identities (created at spawn, deleted on stop),
ephemeral per-pod service identities (self-enrolled with lease), and
persistent runner/app/device identities (service-token provisioned). A
Ziti Management service encapsulates Controller API calls and runs
lease-based garbage collection.

\paragraph{Installation and setup.}
The full stack is packaged as a Helm chart and deployable on any
Kubernetes cluster. A bootstrap repository provides a single-command
setup that provisions the OpenZiti Controller and installs platform
services. Detailed instructions are available in the project
README.

\section{Related Work}

This section reviews existing work on agent orchestration, focusing on closed‑source platforms, open‑source runtimes, and framework‑level isolation, and positions {\tt Agyn} with respect to each.
A systematic comparison of these platforms is provided in Table~\ref{tab:comparison}.

\subsection{Closed Platforms}
AWS Bedrock AgentCore~\cite{bedrock_agentcore} is AWS's managed agent
platform: security and configuration are pieced together from AWS
services (IAM, VPC, Bedrock), not shipped out of the box. Closest to
Agyn but AWS-locked. Anthropic's Claude Managed
Agents~\cite{claudemanagedagents} runs Claude in Anthropic's cloud:
narrow-scoped security features (Vaults, MCP Tunnels, API allowlists)
live inside Anthropic's product, not as defaults.

\subsection{Open-Source Platforms}
Google AX~\cite{google_ax} is a Kubernetes-native distributed agent runtime with A2A
agent integration; the open-source distribution lacks credential
isolation and zero-trust networking.
Platform {\tt kagent}~\cite{kagent} models agents as Kubernetes CRDs with strong
declarative configuration, but runs them as always-on deployments and
mounts model credentials as Pod environment variables.

\subsection{Frameworks and Isolation}
LangGraph~\cite{langgraph}, AutoGen~\cite{wu2023autogen}, and CrewAI are
libraries for \emph{building} agents; {\tt Agyn} operates agents regardless of
framework. IsolateGPT~\cite{isolategpt2025} demonstrates per-tool
isolation at the API level; {\tt Agyn} applies the principle at the container
level. BeyondCorp~\cite{beyondcorp2014} established per-device identity;
OpenZiti~\cite{openziti} extends it across trust boundaries as an
overlay.

\begin{table}[t]
  \centering
  \caption{Comparison of agent operations platforms (symbol ``\ding{51}'' indicates that a feature is present, ``\ding{55}'' indicates it is absent, and ``\textbf{--}'' is for partial support or unknown status).}
  \label{tab:comparison}
  \small
  \setlength{\tabcolsep}{2.5pt}
  \begin{tabular}{lccccccc}
    \toprule
    & \rotatebox{70}{Self-hostable}
    & \rotatebox{70}{Multi-vendor agents}
    & \rotatebox{70}{MCP isolation}
    & \rotatebox{70}{Declarative config}
    & \rotatebox{70}{Serverless}
    & \rotatebox{70}{Credential isolation}
    & \rotatebox{70}{Zero-trust} \\
    \midrule
    \textbf{Agyn}~\cite{Agyn_GitHub}
      & \ding{51} & \ding{51} & \ding{51} & \ding{51} & \ding{51} & \ding{51} & \ding{51} \\
    AgentCore~\cite{bedrock_agentcore}
      & \ding{55} & \ding{51} & \textbf{--} & \textbf{--} & \ding{51} & \textbf{--} & \textbf{--} \\
    Claude Managed Agents~\cite{claudemanagedagents}
      & \ding{55} & \ding{55} & \ding{55} & \textbf{--} & \ding{51} & \textbf{--} & \textbf{--} \\
    Google AX~\cite{google_ax}
      & \ding{51} & \textbf{--} & \ding{55} & \textbf{--} & \ding{51} & \ding{55} & \ding{55} \\
    kagent~\cite{kagent}
      & \ding{51} & \ding{55} & \textbf{--} & \ding{51} & \ding{55} & \ding{55} &  \ding{55} \\
    \bottomrule
  \end{tabular}
\end{table}

\section{Conclusion and Future Work}
We presented Agyn, an open‑source platform for operating LLM‑based agents at organisational scale. Agyn addresses three core requirements that existing  systems fail to satisfy simultaneously: efficient on‑demand execution, agent definition as a code, and zero‑trust access to internal services.

First, Agyn introduces a signal‑driven, stateful serverless runtime that spawns agent instances only when a conversational message arrives and reclaims resources after an idle timeout, while preserving persistent state across cold starts. This combines the elasticity of FaaS with the conversational, stateful nature of agent workloads. Second, Agyn exposes agent definitions and their full harness (MCP servers, secrets, volumes, network identities) through a Terraform provider, enabling version control, peer review, and repeatable deployment – the same governance practices already used for infrastructure. Third, Agyn enforces least privilege through per‑container secret injection, ephemeral x509 identities via OpenZiti, and ReBAC authorisation, ensuring that even a compromised LLM cannot leak credentials or access unauthorised services.

Our evaluation (Table~\ref{tab:comparison}) shows that {\tt Agyn} is the only platform among surveyed systems (AWS Bedrock AgentCore, Claude Managed Agents, Google AX, kagent) that simultaneously provides self‑hosting, pre‑built coding agents such as Claude Code and Codex, MCP servers isolation, agent-as-a-code configuration, serverless execution, credential isolation, and zero‑trust networking.

Limitations remain. Per‑exposure Dial-policy scoping, hard spend-cap enforcement, and extending ReBAC coverage. Ongoing work addresses these gaps.

The platform is released under AGPL‑3.0 and available at 
\url{https://github.com/agynio}. We invite the community to extend the catalogue of agents, and policy patterns that vary across industries.

\subsection*{Acknowledgments}

We extend our deepest gratitude to Rostislav Yavorskiy (Un Minuto Para Saber), our Research and Education advisor. Day after day, Rosti reads our first drafts with the patience of a saint and the red pen of a gentle perfectionist. 

\bibliographystyle{splncs04}
\bibliography{lit}

\end{document}